# ProcessCO v1.3's Terms, Properties, Relationships and Axioms - A Core Ontology for Processes

**Pablo Becker** and **Luis Olsina**
GIDIS_Web, Facultad de Ingeniería, UNLPam, General Pico, LP, Argentina
[beckerp, olsinal]@ing.unlpam.edu.ar

**Abstract.** The present preprint specifies and defines all Terms, Properties, Relationships and Axioms of ProcessCO (*Process Core Ontology*). ProcessCO is an ontology devoted mainly for Work Entities and related terms, which is placed at the core level in the context of a multilayer ontological architecture called FCD-OntoArch (*Foundational, Core, and Domain Ontological Architecture for Sciences*). This is a five-layered ontological architecture, which considers Foundational, Core, Domain and Instance levels, where the domain level is split down in two sub-levels, namely: Top-domain and Low-domain. Ontologies at the same level can be related to each other, except for the foundational level where only ThingFO (*Thing Foundational Ontology*) is found. In addition, ontologies' terms and relationships at lower levels can be semantically enriched by ontologies' terms and relationships from the higher levels. Note that both ThingFO and ontologies at the core level such as ProcessCO, SituationCO, among others, are domain independent with respect to their terms. Stereotypes are the mechanism used for enriching ProcessCO terms mainly from the ThingFO ontology. Note that in the end of this document, we address the ProcessCO vs. ThingFO non-taxonomic relationship verification matrix. Additionally, note that annotations of updates from the previous version (ProcessCO v1.2) to the current one (v1.3) can be found in Appendix A. For instance, 6 axioms were added.

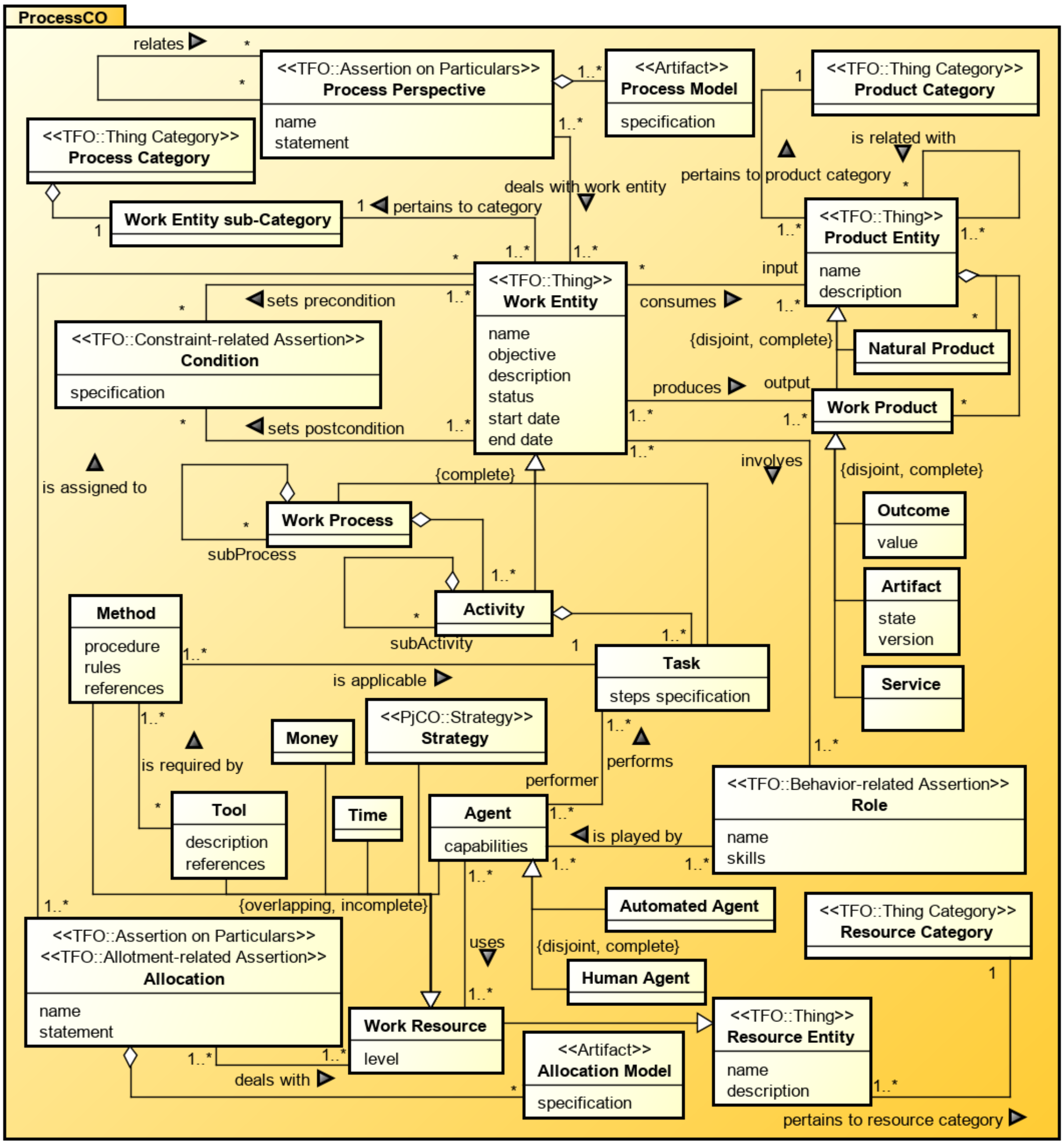

**Figure 1.** ProcessCO v1.3: Core Ontology for work Processes, which is placed at the core level of FCD-OntoArch in Figure 2. This is a revised version of ProcessCO v1.2 [4]. Note that TFO stands for Thing Foundational Ontology [7, 11]. Additionally, note that the ProcessCO v1.0 ontology developed in late 2014 is documented and published in [2]. The new and recent reference [5] contains some enhancements documented in the current preprint; however, in [5], the axioms are not specified.

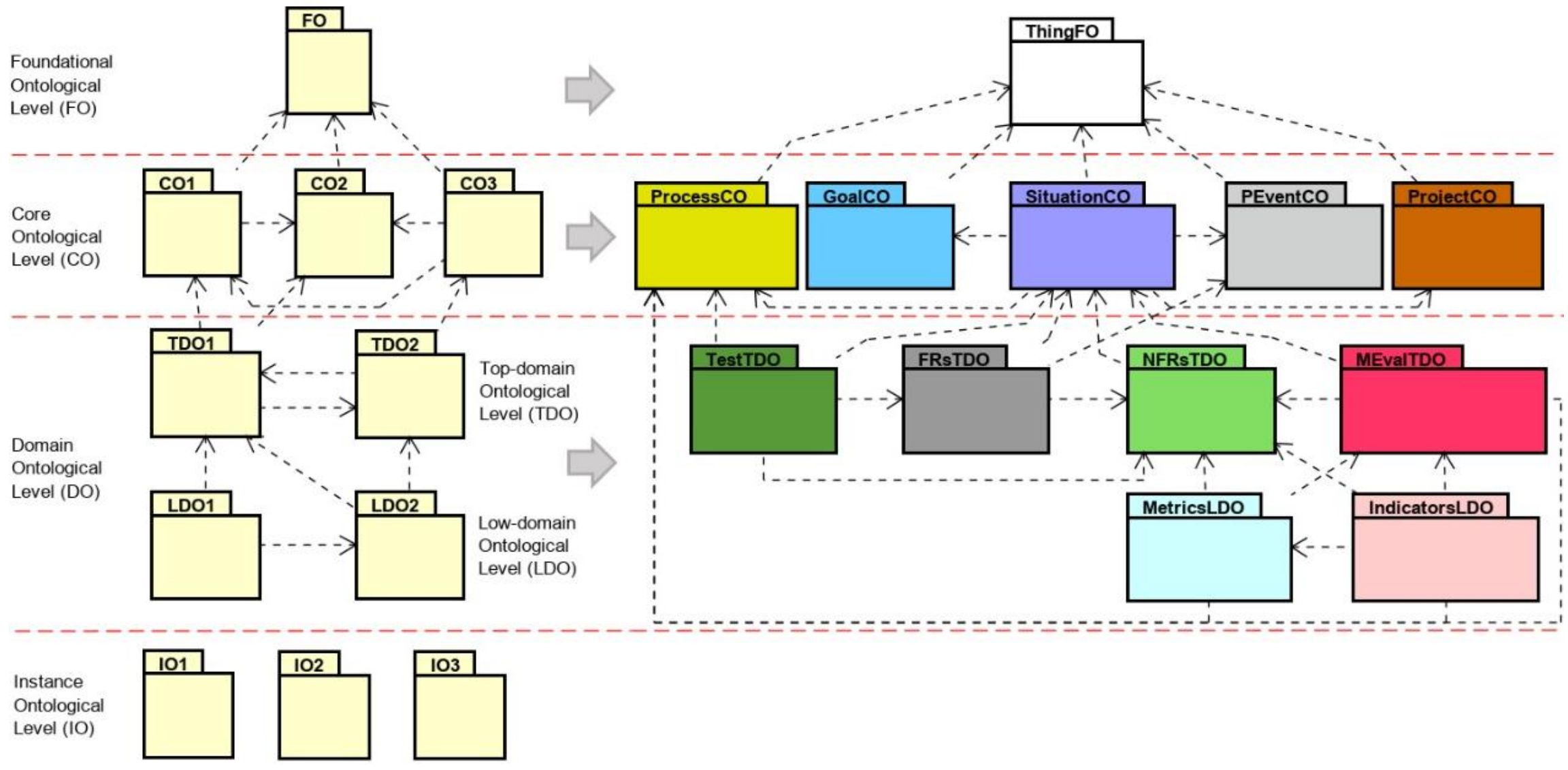

**Figure 2.** Allocating the ProcessCO component or module in the context of the five-layered ontological architecture so-called FCD-OntoArch (*Foundational, Core, and Domain Ontological Architecture for Sciences*) [7].

## Process Component – ProcessCO v1.3's Terms

| Term | Definition |
| --- | --- |
| **Allocation**<br>(synonym: **Allotment)** | It is an Assertion on Particulars, specifically, an Allotment-related Assertion that specifies the assignment of a Work Resource to a Work Entity.<br>Note: For example, a particular Work Resource such as a Method, Tool, Agent, among others, is assigned to a Work Entity in a particular situation. |
| **Allocation Model**<br>(synonym: **Allotment Model)** | It represents an Artifact that specifies and models none or more Allocations of Work Resources.<br>Note 1: An Allocation Model can be seen as an explicit representation of Allocations of Work Resources, which is conceived by a Human Agent with a certain goal purpose.<br>Note 2: An Allocation Model can be modeled by means of informal, semiformal or formal specification languages. |
| **Activity** | It is a Work Entity that is formed by an interrelated set of sub-activities and Tasks.<br>Note 1: A sub-activity is an Activity at a lower granularity level.<br>Note 2: In projects, while Activities are planned, Tasks are scheduled and enacted. |
| **Agent** | It is a Work Resource assigned to a Work Entity to perform a Task in compliance with a Role. |
| **Artifact** | It is a tangible or intangible, versionable Work Product, which can be delivered. |
| **Automated Agent** | It is an Agent, in fact, a non-human Work Resource assigned to a Work Entity, which performs a Task in fulfillment of a Role.<br>Note: In the computing discipline, Automated Agent is |
| | also called bots and it can be, for example, a software agent (i.e., a computer program acting for a user or another program), a robot (i.e., a mechanical device that can perform physical tasks), among others. |
| **Condition** | It is a Constraint-related Assertion that specifies restrictions that must be satisfied or evaluated to true at the beginning (pre-condition) or ending (post-condition) of a Work Entity realization, in given project situations or events. |
| **Human Agent** | It is an Agent, in fact, a human Work Resource assigned to a Work Entity, which performs a Task in fulfillment of a Role.<br><br>Note: A Human Agent embodied by a person –the subject- is the one who conceives goals. |
| **Method** | It is a Work Resource that encompasses the specific and particular way to perform the specified steps in the Work Entity description.<br><br>Note: The specific and particular way of a Method –i.e. *how* the steps in a work description should be made- is represented by a procedure and rules. |
| **Money** | It is a Work Resource that represents a medium of exchange accepted for the payment of goods, services and all kinds of obligations.<br><br>Note: A project requires Money as an asset to be assigned. |
| **Natural Product** | It is a Product Entity that is produced by natural processes.<br><br>Note: A Natural Product is not produced by a Work Entity, however, it can be consumed by a Work Entity. |
| **Outcome** | It is a Work Product that is intangible, storable and processable. |
| **Process Category** | It is a Thing Category (a universal) which has a Work Entity sub-Category.<br><br>Note: Another sub-category may be the Natural Process sub-Category. |
| **Process Model** | It represents an Artifact that specifies and models one or more related Process Perspectives.<br><br>Note 1: A Process Model can be seen as an explicit representation of one or more related views, which is conceived by a Human Agent with a certain goal purpose.<br><br>Note 2: A Process Model can be modeled by means of informal, semiformal or formal specification languages. |
| **Process Perspective**<br><br>(synonym: **Process View)** | It is an Assertion on Particulars that specifies the functional, behavioral, informational, methodological, or organizational view for Work Entities and related concepts.<br><br>Note 1: According to Curtis *et al.* [6], a process can be modeled taking into account four process perspectives: i) *functional* that includes the Work Entities' structure, Work Products as inputs and outputs, etc.; ii) *informational* that includes the structure and interrelationships among Work Products produced or consumed by Work Entities; iii) *behavioral* that models the dynamic view of Work Entities, including Conditions; and, iv) *organizational* that deals with Agents and Roles. Additionally, a *methodological* |
| | perspective is described in [9], which is used to represent the process constructors (i.e., Methods) that can be applied to different descriptions of Work Entities.<br><br>Note 2: Examples of specifications for different Process Perspectives are in [1], and more recently in [3].<br><br>Note 3: A Process Perspective is an Assertion on Particulars, which in turn is also a Behavior-related Assertion, or an Action-related Assertion or any other Assertion type of ThingFO [7, 10, 11]. For example, the informational view is a Structure-related Assertion, the organizational view is an Allotment-related Assertion. |
| **Product Category** | It is a Thing Category (a universal) to which concrete Product Entities belong to. |
| **Product Entity** | It is a Thing (a particular) produced naturally or yielded artificially as a result of a Work Entity. |
| **Resource Category** | It is a Thing Category (a universal) to which concrete Resource Entities belong to. |
| **Resource Entity** | It is a Thing (a particular) that represents an available asset that can be intentionally used for or allocated to something as a means of help, support, or need in a particular event or situation.<br><br>Note: An asset is something with added value for an organization or person. |
| **Role** | It is a Behavior-related Assertion that specifies a set of skills that an Agent must possess in order to perform a Work Entity.<br><br>Note: Skills include abilities, competencies and responsibilities. |
| **Service** | It is a Work Product that is intangible, non-storable and deliverable. |
| **Strategy** | It is a Work Resource that encompasses principles and integrated capabilities such as domain conceptual bases, the specification of process perspectives and methods for helping to achieve a project's goal purpose. |
| **Task** | It is an atomic, fine-grained Work Entity that cannot be decomposed. |
| **Time** | It is a Work Resource that represents a finite, non-storable, perishable and inexorable non-spatial continuum assigned to Work Entities (Processes, Activities and Tasks) when (re-)scheduling their duration in a project.<br><br>Note: See the definition of Time (Context) Entity in SituationCO [8]. |
| **Tool**<br><br>(synonym: **Instrument)** | It is a Work Resource that represents an instrument facilitating the automation and execution of Method procedures and rules.<br><br>Note: An instrument can be physical (hardware), computerized (software) or a combination of both. |
| **Work Entity** | It is a Thing (a particular) that describes the work by means of consumed and produced Work Products, Conditions, and involved Roles. |
| | Note 1: Work Entity represents a Work Process, an Activity or a Task.<br>Note 2: Work Entity represents a Thing conceived by a human being. |
| **Work Entity sub-Category** | A Process's sub-category to which concrete Work Entities belong to. |
| **Work Process**<br>(synonym: **Process**) | It is a coarse-grained Work Entity that is composed of an interrelated set of sub-processes and activities.<br>Note: A sub-process is a Work Process at a lower granularity level. |
| **Work Product** | It is a Product Entity that is consumed or produced by a Work Entity. |
| **Work Resource** | It is a Resource Entity that represents an available asset that can be allotted and assigned to Work Entities.<br>Note 1: Useful Work Resources in any project are at least Agents, Methods and Tools, Strategies, Time, and Money.<br>Note 2: Only scheduled tasks can be performed on a project effectively. |

*Amount of Terms: 30*

## Process Component – ProcessCO v1.3's Attributes or Properties

| Term | Attribute | Definition |
|---|---|---|
| Allocation | **name** | Label or name that identifies the Allocation of Work Resources. |
| | **statement** | An unambiguous textual statement describing the Allocation of Work Resources. |
| Allocation Model | **specification** | The explicit and detailed representation or model of the Allocation in a given language. |
| Agent | **capabilities** | Set of abilities that the Agent has as a performer. |
| Artifact | **state** | State in which the Artifact is. |
| | **version** | Unique identifier, which indicates the level of evolution of the Artifact. |
| Condition | **specification** | Unambiguous specification of constraints, restrictions or circumstances that must be achieved or satisfied. |
| Method | **procedure** | Arranged set of instructions or operations, which specifies how the steps in a Task description must be performed. |
| | **rules** | Set of principles, conditions, heuristics, axioms, etc. associated with the procedure. |
| | **references** | Citation of bibliographical or URL resources, where authoritative and additional information for the Method can be consulted. |
| Outcome | **value** | Numerical or categorical result. |
| Process Model | **specification** | The explicit and detailed representation or model of the Work Entity perspective in a given language. |
| Process Perspective | **name** | Label or name that identifies the Process Perspective.<br><br>Note: Examples of names of process perspectives are: Functional View, Informational View, Behavioral View, Methodological View and Organizational View. |
| | **statement** | An unambiguous textual statement describing the Process Perspective. |
| Product Entity | **name** | Label or name that identifies the Product Entity. |
| | **description** | An unambiguous textual statement describing the Product Entity. |
| Resource Entity | **name** | Label or name that identifies the Resource Entity. |
| | **description** | An unambiguous textual statement describing the Resource Entity. |
| Role | **name** | Label or name that identifies the Role. |
| | **skills** | Set of capabilities, competencies and responsibilities of the Role. |
| Task | **steps specification** | Specification of steps to be followed in order to achieve the Task objective. |
| Tool | **description** | An unambiguous textual statement describing the Tool. |
| | **references** | Citation of bibliographical or URL resources, where authoritative and additional information for the Tool can be consulted. |
| Work Entity | **name** | Label or name that identifies the Work Entity. |
| | **objective** | Aim or end to be reached. |
| | **description** | An unambiguous textual statement describing what to do for achieving the objective of the Work Entity.<br><br>Note: It represents what should be done instead of how it should be performed. |
| | **status** | State in which the Work Entity is. |
| | **start date** | Date or instant of time when the Work Entity starts. |
| | **end date** | Date or instant of time when the Work Entity ends. |
| Work Resource | **level** | Level to which the Work Resource is assigned.<br><br>Note 1: Examples of work level are strategic, project and task. |

| | |
|---|---|
| | Note 2: For example, Time and Money are rather assigned to project level, while Methods, Agents and Tools are rather assigned to task level. |

*Amount of Attributes (Properties): 30*

## Process Component – ProcessCO v1.3's Non-taxonomic Relationships*

| Relationship | Definition |
|---|---|
| **consumes** | In order to achieve its objective, a Work Entity consumes one or more Product Entities. |
| **deals with** | An Allocation deals with one or more Work Resources. |
| **deals with work entity** | A Process Perspective deals with one or more Work Entities. |
| **involves** | A Work Entity involves one or more Roles. In turn, a Role may participate in one or more Work Entities. |
| **is applicable** | A Method is applicable to the description of a Task. In turn, for a Task's description one or several Methods can be applied. |
| **is assigned to** | A scheduled Allocation of Work Resources is assigned to Work Entities for their enactment. |
| **is played by** | A Role is played by one or several Agents. In turn, an Agent plays one or more Roles. |
| **is related with** | A Product Entity is related with none or several Product Entities. |
| **is required by** | A Tool is required by none or several Methods. |
| **performs** | An Agent performs one or more assigned Tasks. In turn, a Task is performed by one or more Agents. |
| **pertains to category** | Work Entities pertain to a Work Entity sub-Category. |
| **pertains to product category** | Product Entities pertain to a Product Category. |
| **pertains to resource category** | Work Resources pertain to a Resource Category. |
| **produces** | A Work Entity produces (modifies, creates) one or more Work Products. |
| **sets postcondition** | A Work Entity may have associated Conditions, which must be accomplished at the end of its realization to be considered finished. <br><br> Note: A post-condition defines any kind of constraint that must evaluate to true before the work described for the Work Entity can be declared completed or finished and on which other Work Entity might depend upon. |
| **sets precondition** | A Work Entity may have associated Conditions, which must be |
| | accomplished before its initiation.<br><br><u>Note:</u> A pre-condition defines any kind of constraint that must evaluate to true before the work described for the Work Entity can start. |
| **relates** | A Process Perspective relates none or several Process Perspectives. |
| **uses** | An Agent uses one or more Work Resources to perform a Task. |

* Note that the ProcessCO non-taxonomic relationships are specialized from ThingFO ones. See below the correspondence between them.

*Amount of non-taxonomic relationships: 18*

# ProcessCO v1.3 vs. ThingFO v1.2 Non-Taxonomic Relationship Verification Matrix

| ProcessCO's Non-taxonomic Relationships | | | | | ThingFO's Non-taxonomic Relationships | | | | |
|---|---|---|---|---|---|---|---|---|---|
| **card** | **Term 1** | **relationship name** | **card** | **Term 2** | **card** | **Term 1** | **relationship name** | **card** | **Term 2** |
| * | Work Entity | consumes | 1..* | Product Entity | 1..* | (Power of) Thing | interacts with other | 1..* | Thing |
| 1..* | Allocation | deals with | 1..* | Work Resource | 1..* | Assertion on Particulars | deals with particulars | 1..* | Thing |
| 1..* | Process Perspective | deals with work entity | 1..* | Work Entity | 1..* | Assertion on Particulars | deals with particulars | 1..* | Thing |
| 1..* | Work Entity | involves | 1..* | Role | * | Thing | defines | * | Assertion |
| 1..* | Method | is applicable | 1 | Task | 1..* | Thing | relates with | 1..* | Thing |
| 1..* | Allocation | is assigned to | * | Work Entity | 1..* | Assertion on Particulars | deals with particulars | 1..* | Thing |
| 1..* | Role | is played by | 1..* | Agent | 1..* | Assertion on Particulars | deals with particulars | 1..* | Thing |
| 1..* | Product Entity | is related with | * | Product Entity | 1..* | Thing | relates with | 1..* | Thing |
| * | Tool | is required by | 1..* | Method | 1..* | (Power of) Thing | interacts with other | 1..* | Thing |
| 1..* | Agent | performs | 1..* | Task | 1..* | (Power of) Thing | interacts with other | 1..* | Thing |
| 1..* | Work Entity | pertains to category | 1 | Work Entity sub-Category | 1..* | Thing | belongs to | * | Thing Category |
| 1..* | Product Entity | pertains to product category | 1 | Product Category | 1..* | Thing | belongs to | * | Thing Category |
| 1..* | Resource Entity | pertains to resource category | 1 | Resource Entity Category | 1..* | Thing | belongs to | * | Thing Category |
| 1..* | Work Entity | produces | 1..* | Work Product | 1..* | (Power of) Thing | interacts with other | 1..* | Thing |
| * | Process Perspective | relates | * | Process Perspective | * | Assertion on Particulars | relates with | * | Assertion on Particulars |
| 1..* | Work Entity | sets postcondition | * | Condition | * | Thing | defines | * | Assertion |
| 1..* | Work Entity | sets precondition | * | Condition | * | Thing | defines | * | Assertion |
| 1..* | Agent | uses | 1..* | Work Resource | 1..* | (Power of) Thing | interacts with other | 1..* | Thing |

# ProcessCO Component - ProcessCO v1.3's Axioms

**A1** description: *For any Work Process that consumes a Product Entity, then this Work Process has at least some subprocess or Activity which consumes the Product Entity.*

**A1** specification:

$$\forall wp_a, \forall pe: \{WorkProcess(wp_a) \land ProductEntity(pe) \land consumes(wp_a, pe) \rightarrow [\exists wp_b: WorkProcess(wp_b) \land consumes(wp_b, pe) \land subProcessOf(wp_b, wp_a)] \lor [\exists a: Activity(a) \land consumes(a, pe) \land partOf(a, wp_a)]\}$$

**A2** description: *For any Activity that consumes a Product Entity, then this Activity has at least some subactivity or Task which consumes the Product Entity.*

**A2** specification:

$$\forall a_1, \forall pe: \{Activity(a_1) \wedge ProductEntity(pe) \wedge consumes(a_1, pe) \rightarrow [\exists a_2: Activity(a_2) \wedge consumes(a_2, pe) \wedge subActivityOf(a_2, a_1)] \vee [\exists t: Task(t) \wedge consumes(t, pe) \wedge partOf(t, a_1)]\}$$

**A3** description: *For any Work Process that produces a Work Product, then this Work Process has at least some subprocess or Activity which produces the Work Product.*

**A3** specification:

$$\forall wproc_a, \forall wprod: \{WorkProcess(wproc_a) \wedge WorkProduct(wprod) \wedge produces(wproc_a, wprod) \rightarrow [\exists wproc_b: WorkProcess(wproc_b) \wedge produces(wproc_b, wprod) \wedge subProcessOf(wproc_b, wproc_a)] \vee [\exists a: Activity(a) \wedge produces(a, wprod) \wedge partOf(a, wproc_a)]\}$$

**A4** description: *For any Activity that produces a Work Product, then this Activity has at least some subactivity or Task which produces the Work Product.*

**A4** specification:

$$\forall a_1, \forall wprod: \{Activity(a_1) \wedge WorkProduct(wprod) \wedge produces(a_1, wprod) \rightarrow [\exists a_2: Activity(a_2) \wedge produces(a_2, wprod) \wedge subActivityOf(a_2, a_1)] \vee [\exists t: Task(t) \wedge produces(t, wprod) \wedge partOf(t, a_1)]\}$$

**A5** description: *If a Work Process involves a Role, then some of its subprocesses or Activities involve it as well.*

**A5** specification:

$$\forall wp_a, \forall r: \{WorkProcess(wp_a) \wedge Role(r) \wedge involves(wp_a, r) \rightarrow [\exists wp_b: WorkProcess(wp_b) \wedge involves(wp_b, r) \wedge subProcessOf(wp_b, wp_a)] \vee [\exists a: Activity(a) \wedge involves(a, r) \wedge partOf(a, wp_a)]\}$$

**A6** description: *If an Activity involves a Role, then some of its subactivities or Tasks involve it as well.*

**A6** specification:

$$\forall a_1, \forall r: \{Activity(a_1) \wedge Role(r) \wedge involves(a_1, r) \rightarrow [\exists a_2: Activity(a_2) \wedge involves(a_2, r) \wedge subActivityOf(a_2, a_1)] \vee [\exists t: Task(t) \wedge involves(t, r) \wedge partOf(t, a_1)]\}$$

**Acknowledgments.** We warmly thank Maria Fernanda Papa and Guido Tebes (both GIDIS_Web research members at the Engineering School, UNLPam, Argentina) for their close collaboration on aspects of the ProcessCO specification.

# Appendix A: Updates from ProcessCO v1.2 to ProcessCO v1.3

Note that the previous version of the ProcessCO ontology (i.e., v1.2) can be found in [4].

- The addition of six axioms, currently labeled A1-A6, which are specified in first-order logic. ProcessCO v1.2 had no specified axiom.
- The addition of new constraints expressed in the diagram with the labels {`disjoint`, `complete`} and {`overlapping`, `incomplete`}, as depicted in Fig. 1.
- The addition of the ProcessCO v1.3 vs. ThingFO v1.2 non-taxonomic relationship verification matrix. As a consequence, many non-taxonomic relationships were renamed in order to reflect the reuse by refinement of the corresponding ThingFO relationships.
- The addition in the stereotypes of the namespaces corresponding to each ontology of FCD-OntoArch, which semantically enriches each term.
- The addition of 2 new references, namely: [5] and [11]. Some references have been updated since preprints were upgraded, for example, [8] and [10].